\documentclass[conference]{IEEEtran}
\IEEEoverridecommandlockouts

\usepackage{cite}
\usepackage{amsmath,amssymb,amsfonts}
\usepackage{algorithmic}
\usepackage{algorithm}
\usepackage{enumerate}
\usepackage{array}
\usepackage{graphicx}
\usepackage{textcomp}
\usepackage{xcolor}
\usepackage{multirow} 
\usepackage{booktabs} 
\def\BibTeX{{\rm B\kern-.05em{\sc i\kern-.025em b}\kern-.08em
    T\kern-.1667em\lower.7ex\hbox{E}\kern-.125emX}}
\begin{document}

\title{Measuring Human Involvement in AI-Generated Text: A Case Study on Academic Writing
}

\author{
    Yuchen Guo\textsuperscript{1,2}, 
 Zhicheng Dou\textsuperscript{1,2}, Huy H. Nguyen \textsuperscript{2}, Ching-Chun Chang\textsuperscript{2}, Saku Sugawara\textsuperscript{2}, Isao Echizen\textsuperscript{1,2‡}\\
    \textsuperscript{1}\textit{Dept. of Information \& Communication Engineering (ICE), The University of Tokyo (Japan)} \\
    \textsuperscript{2}\textit{National Institute of Informatics (Japan)} \\
    \textit{Email: \{guoyuchen, dou, nhhuy, ccchang, saku, iechizen\}@nii.ac.jp}
    \thanks{\textsuperscript{‡}Corresponding author}
}

\maketitle

\begin{abstract}
Content creation has dramatically progressed with the rapid advancement of large language models like ChatGPT and Claude. While this progress has greatly enhanced various aspects of life and work, it has also negatively affected certain areas of society. A recent survey revealed that nearly 30\% of college students use generative AI to help write academic papers and reports. Most countermeasures treat the detection of AI-generated text as a binary classification task and thus lack robustness. This approach overlooks human involvement in the generation of content even though human-machine collaboration is becoming mainstream. Besides generating entire texts, people may use machines to complete or revise texts. Such human involvement varies case by case, which makes binary classification a less than satisfactory approach. We refer to this situation as \textit{participation detection obfuscation}. We propose using \textit{BERTScore} as a metric to measure human involvement in the generation process and a \textit{multi-task RoBERTa-based regressor} trained on a token classification task to address this problem. To evaluate the effectiveness of this approach, we simulated academic-based scenarios and created a continuous dataset reflecting various levels of human involvement. All of the existing detectors we examined failed to detect the level of human involvement on this dataset. Our method, however, succeeded (F1 score of 0.9423 and a regressor mean squared error of 0.004). Moreover, it demonstrated some generalizability across generative models. Our code is available at https://github.com/gyc-nii/CAS-CS-and-dual-head-detector
\end{abstract}

\section{Introduction}
Large language models (LLMs) like ChatGPT have advanced natural language processing by utilizing vast amounts of text data to understand and generate human-like language. These models are used in various applications, such as chatbots and content creation, enhancing tasks like research assistance, coding, and translation \cite{bahrini2023chatgpt, achiam2023gpt}. However, the potential misuse of LLM-generated content raises grave concerns. This is particularly true in academia given that high-quality AI-generated responses can facilitate academic dishonesty by enabling individuals to produce credible work without genuine research \cite{das2024security}\cite{news_A}. Research indicates that 69\% of U.S. universities have policies on the use of AI \cite{university_policy}. While it is hard for people to judge whether text is AI generated \cite{news_B}, a survey revealed that nearly 30\% of college students used ChatGPT for schoolwork\cite{kyaw2023survey}, highlighting the need for robust methods to address AI-assisted academic misconduct.

\begin{figure}[t]
\centering
\includegraphics[width=\linewidth]{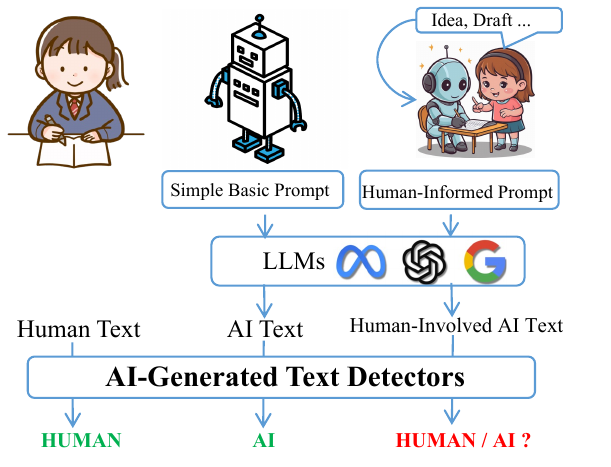} 
\caption{Text can be entirely authored by a person, generated by an LLM using simple basic prompts, or created through a collaborative process involving human-informed prompts, which incorporate human input. This mixed-origin text can confound existing detection mechanisms, leading to the participation detection obfuscation problem.}
\label{fig:background}
\end{figure}

Several binary classification countermeasures have been established to address the problem of academic cheating. The challenge, however, lies in the ambiguity surrounding the definition and detection of AI-assisted academic cheating. Many universities have established guidelines on the permissible use of AI tools in academic work, but their enforcement remains inconsistent. Most institutions allow the use of AI for supportive tasks such as idea generation and grammar improvement but consider the use of unmodified AI-generated content as academic misconduct \cite{gulumbe2024balancing, chan2023comprehensive}. Binary classification detectors, however, struggle to accurately identify AI-generated text or precisely quantify the human contribution in scenarios involving human-AI collaboration \cite{yu2023cheat, dou2024enhancing}. This leads to the participation detection obfuscation problem, as illustrated in Fig.~\ref{fig:background}. For example, a person may use an LLM to help produce an idea or draft (defined as \textit{human information}) and then input it to an LLM via a prompt to generate text without subsequent human modification.

The above research has demonstrated that the human-likeness of LLMs improves with \textit{the amount of human-provided information included in the prompts} input to LLMs for generating text, which we quantify as \textit{human involvement} in the generated text. As human involvement increases, the resulting texts become increasingly indistinguishable from those written solely by a person. The traditional dichotomous approach to detection categorizes text as human or AI-generated and thus lacks the flexibility needed to provide educators with insightful explanations for the detection results. Thus, there is an urgent need for advanced tools to accurately quantify the extent of human involvement in academic work based on prompts and generated text, followed by a model that can estimate human involvement without prior knowledge of the prompt.

To address the need for quantification, we propose using the BERTScore \cite{zhangbertscore} as a metric to measure human involvement in academic texts. This metric, normalized to a range of 0 to 1 using max-min mapping, provides a continuous scale of human involvement. A higher score means a higher level of human involvement in the generated text. To better represent actual scenarios, we created the Continuous Academic Set in Computer Science (CAS-CS) dataset by varying the degree of human involvement in the prompts provided to the LLM. Such a dataset should confuse binary classification detectors. 

To estimate human involvement in generated content, we propose using a RoBERTa-based regression model as a detector. We developed an interpretability module that integrates with the regressor to identify specific words contributed by a person, thus enhancing the explanation of the results. We used the CAS-CS dataset to evaluate the performance of our proposed detector. Our experimental results indicate that it outperforms existing detectors on both classification and regression tasks, for both traditional binary datasets and the CAS-CS dataset. The interpretability module also shows promising performance, suggesting that a method based on our approach can reliably assess human involvement in academic texts.

The contributions of our work are threefold:
\begin{itemize}
\item We have demonstrated a novel usage of the BERTScore as a continuous-scale metric to quantify the extent of human involvement in AI-assisted writing between prompt entry and text generation.
\item We have constructed a comprehensive dataset (the CAS-CS dataset) of ChatGPT-generated articles with varying levels of human input that corresponds to real-world situations more realistically than existing datasets.
\item We have devised a RoBERTa-based regression detector with an interpretability module that estimates the degree of human involvement in written content without knowing the prompt beforehand and that is able to generalize across different generative models. 
\end{itemize}
\section{Related work}
Statistical methods have been the focus in previous research on LLM detectors. With the evolution of generative models, statistical content has gradually been made more complicated to ensure the accuracy of LLM detectors. Gallé et al. \cite{galle2021unsupervised} classified text by counting the number of n-grams that appear multiple times in the text. Hamed and Wu \cite{hamed2024detection} classified text by the frequency of bigram occurrence.

Such statistical methods can no longer achieve stable performance on the latest generation models. Therefore, researchers use existing generation AI models to analyze the statistical features further for classification. Verma et al. \cite{verma2024ghostbuster} classified text by examining the generation probabilities of a series of generative models. Mao et al. \cite{mao2024raidar} used a generative model to complete the writing of the text to be tested and then used statistical methods to analyze the similarity between the two for classification. Both studies used deep neural networks, demonstrating that deep learning is the primary research and development direction for detectors.

In studies on using deep learning for classification, some researchers extracted the features first and then input them into a deep neural network for classification. Some researchers \cite{mindner2023classification,li2023origin} extracted the language features and the contrast features and then input them into a neural network for classification. Other researchers directly used pre-trained deep neural network models and input complete texts for classification, which is the most common method.  A number of researchers demonstrated that a pre-trained BERT model can achieve excellent performance in various fields with only small training sets \cite{guo2023close,liu2023argugpt,chen2023gpt,wang2023implementing}. Several researchers solved the problem of insufficient training data by using contrastive learning \cite{yan2021consert,gao2021simcse,chen2022dual}.

However, the existing classification criteria used by classifiers trained on deep neural networks depend entirely on the training dataset's ground truth: the definitions of human text and AI text in the dataset. Although there is a consensus on the definition of human text in current research, the definition of AI text depends on the prompt, so a binary ground truth does not fit all real-world scenarios. AI-generated texts can be divided into three categories: 'generation abstracts, polish abstracts, and mix abstracts' \cite{yu2023cheat} or 'GPT-written full abstracts, GPT-completed abstracts, and  GPT-polished abstracts' \cite{liu2023detectability}. Dou et al. \cite{dou2024enhancing} analyzed the effect of human information in the prompt on the generated text. Moreover, binary classification does not work well in situations where the ground truth changes depending on the prompt. Therefore, a new detection system is needed that can provide users more information than simply the classification results.

\begin{figure}[htbp]
\centering
\includegraphics[width=\linewidth]{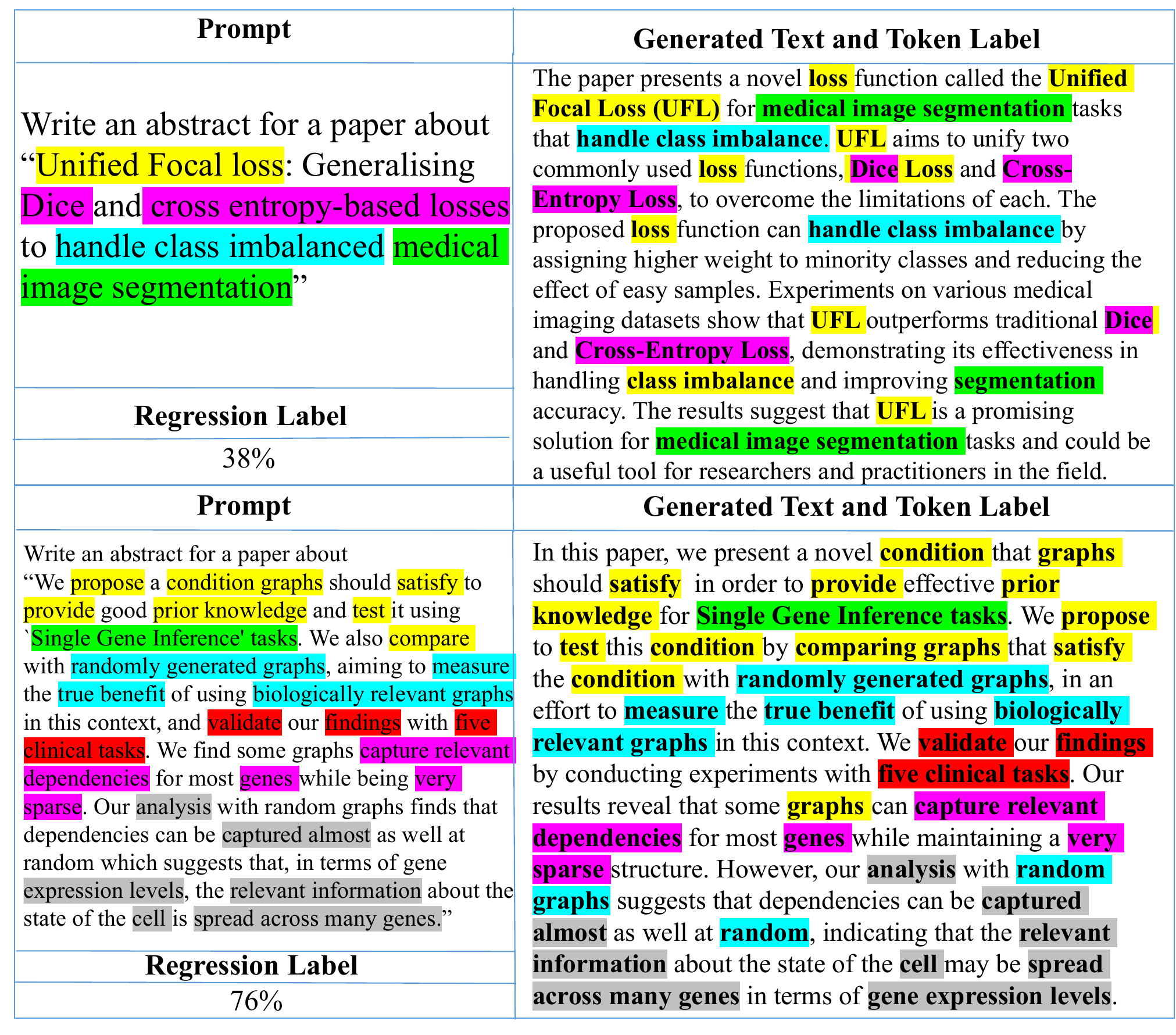} 
\caption{Example text reflecting low and high human involvement. The same tokens are marked in the same color.}
\label{fig:example}
\end{figure}

\section{Problem Statement}
\subsection{Application Scenario}
The target application is the provision of an accurate and robust way to measure the percentage of human involvement in written text (e.g., research papers, technical reports, school assignments), especially for collaborative texts that traditional detectors cannot handle well. For example, a test taker may use prompts that include their own ideas or thoughts to produce a collaborative AI-generated essay. Our envisioned system would give the examiner a score that reveals the degree of human involvement in the essay. The examiner can also use an interpretive tool to detect which words were provided by the test taker, as illustrated in Fig.~\ref{fig:example}. The prompts provided to the LLM are shown on the left, and the generated text is shown on the right. The task was to identify the human contribution (left side) by using regression and to identify the human-contributed words (highlighted on the right). More examples can be found in the supplementary materials.

\subsection{Problem Formulation}
Our approach to measuring human involvement in AI-generated text, particularly in the context of academic writing, involves two main tasks: regression and classification. Given an input prompt \( P \) and the text \( G \) generated by an LLM, we proceed as follows:

First, we compute the recall score between \( P \) and \( G \) using BERTScore to reflect how much \( P \) is involved in \( G \). This recall score is then normalized using min-max normalization to obtain regression label \( y_{\text{reg}} \), which represents the human involvement in the generated text. 

\begin{equation}
y_{\text{reg}} = \frac{\text{BERTScore}(P, G) - \min(\text{BERTScore})}{\max(\text{BERTScore}) - \min(\text{BERTScore})}
\end{equation}

Next, we identify the common tokens between \( P \) and \( G \) and use whether the token after lemmatization is the same or not as the classification label \( y_{\text{cls}} \). \( y_{\text{cls}} \) is a 1×368 vector, where each value is either 0 or 1: 0 means the token is not in \( P \) and is a stop word, punctuation, or padding; 1 means the token is in \( P \). This process can be described by

\begin{equation}
y_{\text{cls}} = \text{CommonTokens}(P, G),
\end{equation}

where \(\text{CommonTokens}(P, G)\) denotes tokens that appear in both \( P \) and \( G \) except stop words and punctuation.

We train a dual-head model to estimate human involvement. During training, the model uses both \( P \) and \( G \) to learn from the regression and classification labels while during estimation, the model requires only the generated text \( G \) as input. The first head of the model estimates human involvement (regression output), and the second head estimates if each token is from a person (classification output). The estimated outputs are represented as \(\hat{y}_{\text{reg}}\) and \(\hat{y}_{\text{cls}}\), respectively:

\begin{equation}
\hat{y}_{\text{reg}} = f_{\text{reg}}(G)
\end{equation}
\begin{equation}
\hat{y}_{\text{cls}} = f_{\text{cls}}(G)
\end{equation}

This dual-head model enables simultaneous evaluation of the human involvement and the specific human-contributed words in the AI-generated text, thus providing a comprehensive measure of human contribution.
\section{Methodology}
\subsection{BERTScore-based Metrics}
\subsubsection{BERTScore Definitions}

BERTScore is a metric designed to evaluate the similarity of a candidate text to reference texts using BERT embeddings. A detailed description can be found in the supplementary materials. 
 
BERTScore has three values:

\begin{itemize}
\item \textbf{Precision}: Precision measures how many of the embeddings in the candidate text can be found in the reference text. It is calculated by finding the closest reference token for each token in the candidate text and averaging their scores.
\item \textbf{Recall}: Recall measures how many of the embeddings in the reference text can be found in the candidate text. It is calculated by finding the closest candidate token for each token in the reference text and averaging their scores. 
\item \textbf{F1-score}: The F1-score is the harmonic mean of precision and recall; it provides a balanced measure that considers both the inclusion and coverage of information. It combines precision and recall to give a single score that balances the trade-off between them.
\end{itemize}

\subsubsection{BERTScore for AI-Generated Text Detection}
\label{sec:bert_for_AI}
In the context of LLMs, prompts are essential inputs that guide the LLM generation of content. The prompt represents the user's contribution to the creation process while the generated text is the output produced by the LLM on the basis of the prompt. BERTScore serves as a robust tool for assessing the semantic similarity between the prompt and the generated text. Since BERTScore values typically range between 0.3 and 1, we perform a max-min mapping to normalize them as `utilization,' `human involvement,' and `similarity' over the range 0 to 1. This normalization better aligns with the real-world scenarios we aim to evaluate. The three normalized values can be interpreted as follows:

\begin{itemize}
\item \textbf{Precision (Utilization)}: This measures how much information in the generated text matches the information in the prompt. It reflects the extent to which the LLM utilized the human-contributed content. High precision indicates that the generated text closely follows the input provided by the user.
\item \textbf{Recall (Human involvement)}: This measures how much information in the prompt appears in the generated text. It signifies the degree of human involvement in the output. This value is critical here as it highlights the human contribution to the generation process.
\item \textbf{F1-score (Similarity)}: This represents the overall similarity between the prompt and the generated text, balancing both precision and recall. It provides a holistic measure of how closely the generated text aligns with the prompt regarding content and context.
\end{itemize}

\begin{figure}[htbp]
\centering
\includegraphics[width=\linewidth]{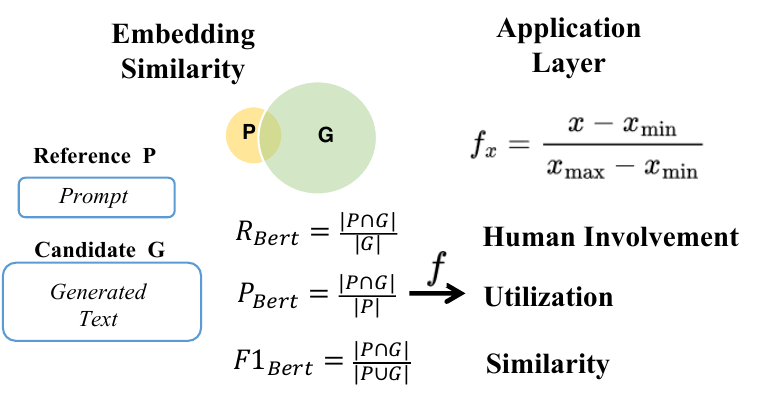} 
\caption{Relationships between metrics.}
\label{fig:score}
\end{figure}

The specific relationships between these metrics are illustrated in Fig.~\ref{fig:score}.
\\
To verify the credibility of BERTScore-based label, we also performed human judgment evaluation, as described in Sec~\ref{sec:H_E}.

\subsection{Continuous CAS-CS Dataset, Polarized Dataset PAS-CS}
Due to text length limitations, we utilized abstracts from actual papers to construct our dataset. Several existing datasets, mostly polarized datasets such as the CHEAT dataset, consider human involvement but often use a discrete form of prompts. For example, the prompts might be "generate an abstract with the title [real title]" or "generate an abstract about [entire actual paper]". This approach does not accurately reflect real-world application scenarios in which the human contribution can range from 0\% to 100\% of the generated text and does not simply specify the topic or the entire content. To address this limitation, we used a random sampling algorithm to create our continuous dataset.

\begin{figure}[htbp]
\centering
\includegraphics[width=\linewidth]{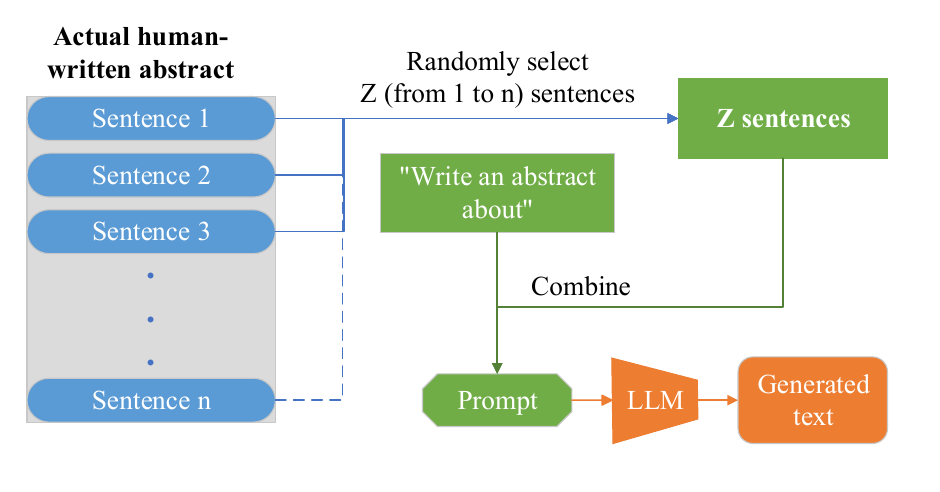} 
\caption{Method for generating continuous dataset.}
\label{fig:dataset}
\end{figure}

As shown in Fig.~\ref{fig:dataset} (the blue areas), we begin with actual human-written abstracts, each consisting of n sentences. A random number Z is generated within the range 1 to n. Next, Z sentences are chosen and combined with "Write an abstract on the basis of" to create a prompt (the green areas), which is sent to the LLM to generate the text (the orange areas). By using this method, we can generate diverse texts encompassing varying degrees of human involvement and thereby create the CAS-CS dataset. This method enables the computation of various BERTScores between the actual abstract content and the generated text, providing metrics for subsequent analysis. Moreover, it enables identification of the tokens in the generated text that were human contributions on the basis of the pairs of the prompt and the generated text, that is, which tokens in the generated text appear in the prompt.

Additionally, we created a polarized dataset comprising texts generated with minimal human involvement and actual human-written abstracts, designated as the Polarized Academic Set in Computer Science (PAS-CS). Using these methods, we can simulate a wide range of potential scenarios when the sample size is sufficiently large. 

\subsection{RoBERTa-based Dual-head Model}
RoBERTa \cite{liu2019roberta} is an advanced variant of BERT \cite{devlin2019bert} known for its exceptional feature extraction capabilities in natural language processing tasks. RoBERTa outputs a 768-dimensional vector for each token in the input sequence. This high-dimensional representation encapsulates the rich contextual information and semantic nuances learned by the model during training. Each dimension in the vector corresponds to a unique aspect of the token's meaning, capturing syntactic, lexical, and contextual features derived from the extensive training data. These vectors, known as embeddings, serve as powerful feature representations for downstream tasks.

\begin{figure*}[t]
\centering
\includegraphics[width=\linewidth]{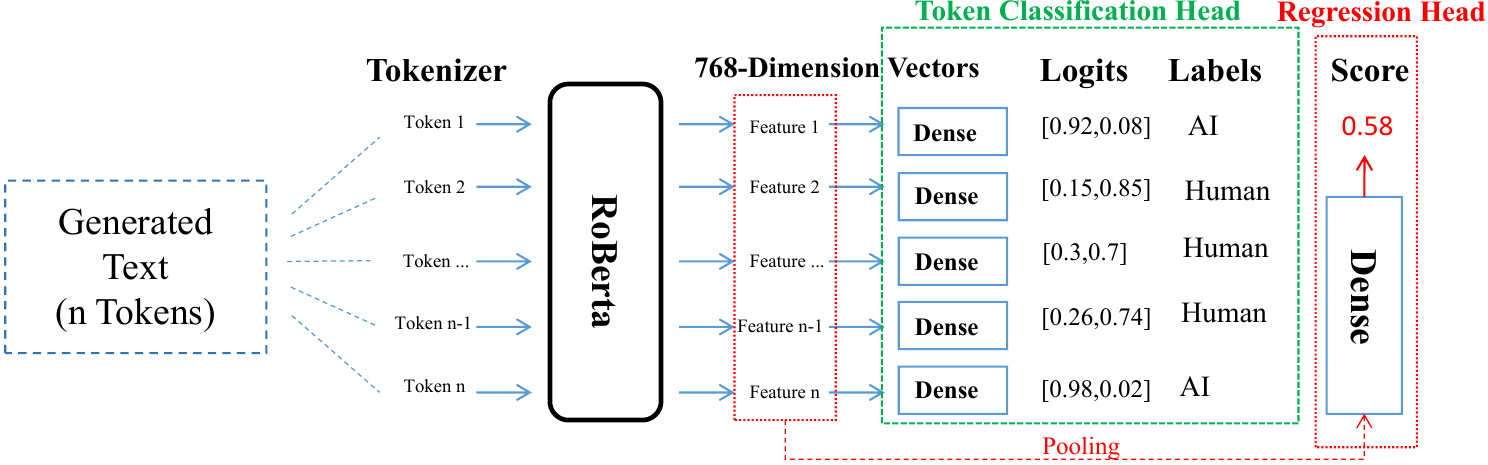} 
\caption{Structure of dual-head model.}
\label{fig:model}
\end{figure*}

We used a shared RoBERTa encoder to create a two-headed model to solve the two downstream tasks. The first regression head is designed to determine the degree of human involvement in the generation of the texts. The second token classification head uses the same encoder to complete the token classification task, aimed at identifying the human-contributed words to better explain the value of human involvement. The model's structure is illustrated in Fig.~\ref{fig:model}.
\subsubsection{Token Classification Head}
In a typical binary classification task using RoBERTa, the model processes input text to generate contextualized token embeddings, each represented as a 768-dimensional vector. To enhance interpretability, word removal and lemmatization are performed to ensure that these words are not labeled. The tokens remaining in the generated text that were present in the original prompt are labeled in accordance with RoBERTa's word segmentation. The logits of each token are then sent to the regression head.
 
\subsubsection{Regression Head}
The token embeddings are subjected to a pooling operation, commonly extracting the embedding of the [CLS] token (the first token of the sequence), designed to capture the aggregate representation of the entire input sequence. This pooled representation is subsequently passed through a fully connected layer, which outputs the probability scores for the two classes(AI-generated and human-contributed).

In our approach, we adapt this process to estimate human involvement by modifying the output layer to produce a single-dimensional output representing the degree of human involvement. All other parameters remain unchanged to ensure a relatively fair comparison with other detectors.

\subsection{Evaluation Method}
\label{sec:BST}

Previous studies predominantly focused on polarized datasets and binary classification detectors to compare the performance. We used the \textit{binarization static threshold (BST)} to divide the ground truth between machine-generated and human-contributed text. The labels indicate the degree of human contribution. When regression label \( y_{\text{reg}} \) in the CAS-CS dataset is less than the \textit{binarization static threshold (BST)}, ground truth \textit{G} is mapped to machine generation. Conversely, when \( y_{\text{reg}} \) exceeds the BST, the ground truth is mapped to the human contribution, as shown in (5). Likewise, for the regression detector, if output \( \hat{y}_{\text{reg}} \) exceeds the BST, estimation P is mapped to human contribution.
\begin{equation}
G, P = 
\begin{cases} 
\text{Human contribution}, & \text{if } y_{\text{reg}},\hat{y}_{\text{reg}} > \text{BST},
\\
\text{AI generation}, & \text{if } y_{\text{reg}},\hat{y}_{\text{reg}} \leq \text{BST}.
\end{cases}
\end{equation}
Fixing BST changes the CAS-CS dataset into a polarized dataset. Furthermore, the estimation detector results will fall into two classes. Given these considerations, we will be able to conduct two experiments. First, we can test an existing binary classifier on the CAS-CS dataset with the BST fixed. We then can convert our regression detector into a binary classifier and test it on polarized dataset benchmarks with the BST set. At the same time, we can also identify the practical significance of BST, which is that instructor may set different requirements for submitted documents (e.g., considering more than 20\% as cheating).
\section{Experiment}
\subsection{Experiment Setup}
\subsubsection{Datasets}
To validate the effectiveness of our proposed  methods, we conducted the experiments using five datasets:

\begin{itemize}
\item\textbf{Continuous CAS-CS Dataset.} Using ChatGPT, we collected 55,000 human-AI collaboration-generated texts in the computer science domain, with human involvement ranging from 0 to 1. Additionally, we collected 1,000 human-AI collaboration-generated texts from Claude, Gemini, GPT-4, and Falcon to test the generalizability of the model used in our proposed method across different AI systems.

\item\textbf{Polarized Dataset PAS-CS.} We collected 2,000 generated abstracts with minimal human involvement and 2,000 human-authored abstracts to enable a fair comparison with existing detectors.

\item\textbf{CHEAT Benchmark} The CHEAT public dataset \cite{yu2023cheat} uses ChatGPT with the prompt ``Generate a 200-word abstract of the input paper in English based on its title and keywords," we collected 15,000 computer science abstracts from the ``Generate'' subset from their public dataset.

\item\textbf{GPA Benchmark} We collected 10k computer science abstracts from the GPABench2 public dataset \cite{liu2023detectability}, which contains 600k GPT-written texts created using ChatGPT and prompts common to academic writing, and an equal number of human-authored abstracts to enable a fair comparison.

\item\textbf{Sa Benchmark.} We collected 1000 of the 3000 texts in the Sa benchmark public dataset \cite{mosca2023distinguishing} generated by the  model ChatGPT as the minimum human involvement generated text. The collected texts were from various academic fields and were not limited to computer science. We also collected an equal number of human-authored abstracts to enable a fair comparison.
\end{itemize}
\subsubsection{Baseline Detectors}
Besides our regression detector, we used four existing binary classification detectors for detecting generated texts: OpenAI detector \cite{solaiman2019release}, ChatGPT detector \cite{guo2023close}, Academic detector \cite{sivesind_2023}, and DetectGPT detector \cite{mitchell2023detectgpt}. We compared their performances on the continuous dataset and polarized dataset with that of our proposed regressor.

\subsubsection{Evaluation metrics}
For the regression task, we used the MSE as the primary evaluation metric, supplemented with the accuracy of label errors within 0.15. For binary classification comparisons, we used the accuracy (ACC) and area under the curve (AUC) metrics. For token classification, we used the accuracy (ACC) and F1 scores to evaluate performance. 

\subsubsection{Training Details and Parameters}
We selected ``en-sci" as the backbone when using BERTScore to label the dataset. For the dual-head RoBERTa-based model, the training was conducted with a learning rate of 1e-6, a batch size of 64, and 100 epochs. The model was trained with L2 regularization with parameter set to 0.001. The loss function (6) consisted of MSE loss (7) for measuring regression loss and cross-entropy loss (8) for measuring token classification loss. To address the imbalance in token labels, we added weights to the cross-entropy loss function of 1.0 and 1.2 for classes 0 and 1, respectively.
\begin{align}
\mathcal{L} &= \mathcal{L}_{\text{MSE}} + \mathcal{L}_{\text{CE}}\\
\mathcal{L}_{\text{MSE}} &= \text{MSE}(y_{\text{reg}}, \hat{y}_{\text{reg}}) \\
\mathcal{L}_{\text{CE}} &= \sum_{i=1}^{N} \left[ 1.0 \cdot \text{CE}_0(y_{\text{cls},i}, \hat{y}_{ \text{cls},i}) + 1.2 \cdot \text{CE}_1(y_{\text{cls},i}, \hat{y}_{\text{cls},i}) \right] 
\end{align}

\subsection{Regression Label Credibility}
\label{sec:H_E}
When using an AI detector to determine whether student submissions were, to some degree, AI-generated, a teacher is the ultimate judge of the degree of human involvement. Since teachers are human, they depend on the detector results when making that judgment. Therefore, to maximize the credibility of the detector results and reduce users' concerns about reliability, we designed a human evaluation experiment to ascertain whether our definition of human involvement in Sec.~\ref{sec:bert_for_AI} is consistent with human judgment. We describe the implementation settings in detail at  https://github.com/gyc-nii/human-evaluation.

\begin{figure}[t]
\centering
\includegraphics[width=\linewidth]{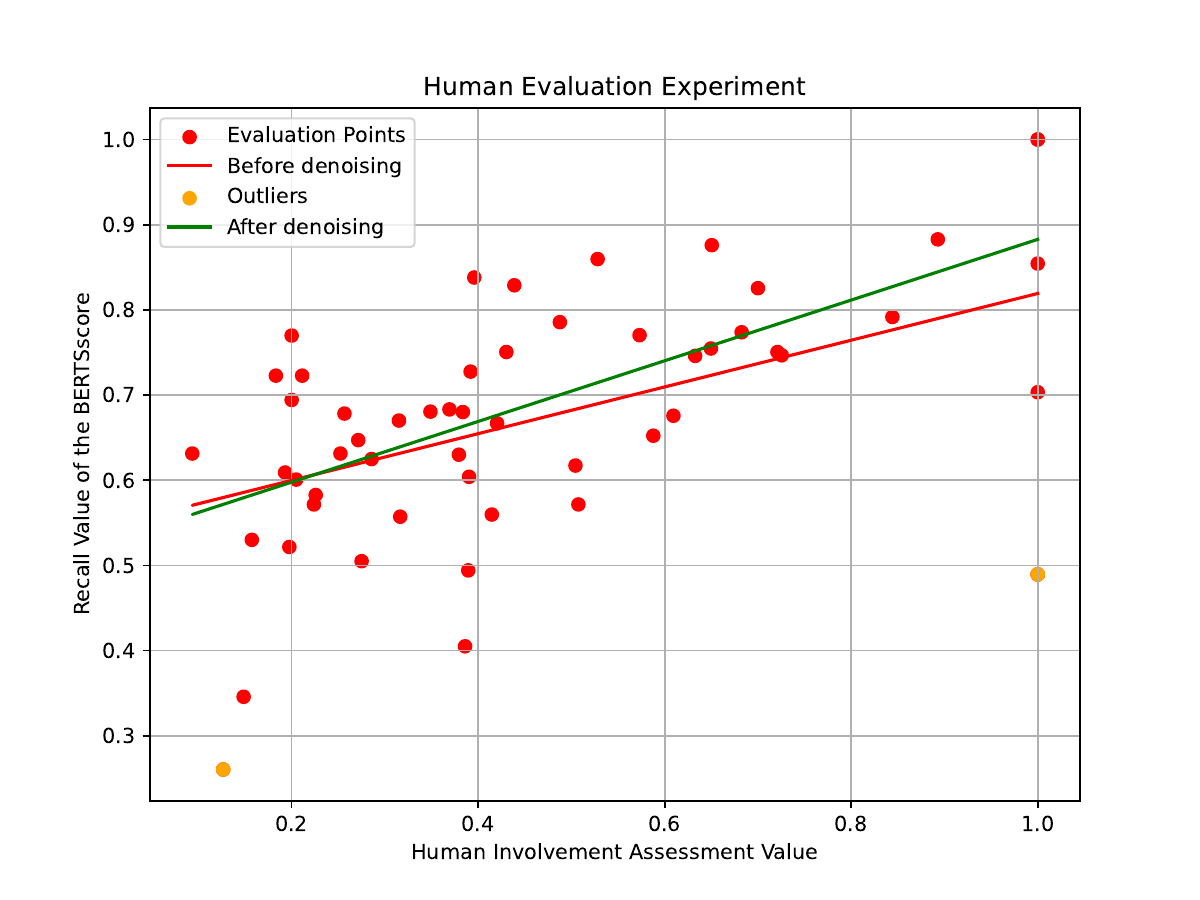} 
\caption{Results of human evaluation. Spearman rank correlation coefficients were 0.52 before denoising and 0.64 after denoising.}
\label{fig:h_e}
\end{figure}

In our human evaluation experiment, we had ten testers (aged 19 to 43; five men and five women) assess the degree of human involvement in the generation of 55 texts. We also calculated the recall value of the BERTScore for the same texts. The results are plotted in Fig.~\ref{fig:h_e}. The ordinate shows the recall value of the BERTScore, which is the regression label we propose using for measuring human involvement. The horizontal axis shows the tester-assessed human involvement values, which were obtained by dividing the number of words identified by the tester as human contributed by the total number of words in the text. For more details, please see our experiment link, which contains the detailed instructions for the experiment. 

Using the original data, we obtained a Spearman rank correlation coefficient of 0.52 between the values of our regression label and the human evaluation values. Considering that people may make mistakes, we removed the outliers (values with residuals greater than two standard deviations). The final Spearman rank correlation coefficient after denoising was 0.64. To facilitate trend observation, we drew two fitting lines.

The experimental results demonstrate to some extent that our regression label is significantly positively correlated with human judgment. This supports the credibility of our regression label and shows that the dataset we designed on the basis of the labels, including the trained model, is consistent with human evaluation of human involvement.

\subsection{Dual-head Model Performance}
To assess the benefits of the dual-head model, we conducted an ablation study. We trained a single-head regression model and a single-head token classification model using the same training set and parameters and tested their performance on the same validation set.
\begin{table}[htbp]
\caption{Ablation study on the dual-head model}
\centering
\begin{tabular}{lccc}
\hline
\textbf{MODEL} & \textbf{MSE} & \textbf{Token ACC} & \textbf{Token F1 } \\
\hline
Reg & 0.006 & - & - \\
Token & - & 0.9485 & 0.9387 \\
Reg+Token & 0.004 & 0.9514 & 0.9423 \\
\hline
\end{tabular}
\label{tab:Ablation Study}
\end{table}

The results (Table~\ref{tab:Ablation Study}) demonstrate that integrating two downstream tasks within the same encoder benefits both tasks. Identifying which words are human-generated aids the regression head in better estimating human involvement, and vice versa. These findings also indicate that the token classification task performs well, with a high probability of accurately identifying human-generated tokens in the text, which improves the interpretability of human involvement results.

\subsection{Evaluation on Polarized Datasets}
To ensure a fair comparison with existing binary classification detectors, we converted our detector into a binary classifier by setting the BST threshold to 0.5 and conducted experiments on various polarized dataset benchmarks. As previously described, these datasets comprise texts generated with minimal human involvement and abstracts that are purely human-authored. The results (Table~\ref{tab:Polarized}) indicate that while the accuracies of all the detectors were suboptimal, the AUC scores were relatively satisfactory. This suggests that although the binary classifiers possess the ability to distinguish between the two types of text, their accuracy is hindered by the selection of the training set. The classifiers tend to perceive texts with human involvement as more similar to human-authored texts although they remain somewhat distinct from genuine human texts. Thus, we conducted an analysis of the classifiers and regressors to understand these observations better.

\begin{table*}[!ht]
\caption{Accuracy and AUC of existing detectors and our regression detector(binarization static threshold=0.5) on polarized dataset PAS-CS(ChatGPT) and other polarized benchmarks. * results from paper \cite{yu2023cheat}}
\centering
\begin{tabular}{lcccccccccc}
\hline
\multirow{2}{*}{Dataset/Detector} & \multicolumn{2}{c}{Openai-D} & \multicolumn{2}{c}{ChatGPT-D} & \multicolumn{2}{c}{Academic-D} & \multicolumn{2}{c}{DetectGPT} & \multicolumn{2}{c}{\begin{tabular}[c]{@{}c@{}}Regression model(ours) \\BST = 0.5\end{tabular}} \\
\cline{2-11}
 & ACC & AUC & ACC & AUC & ACC & AUC & ACC & AUC & ACC & AUC \\
\hline
PAS-CS & 0.60 & 0.69 & 0.70 & 1.00 & 0.61 & 0.92 & 0.51 & 0.94 & \textbf{1.00} & \textbf{1.00} \\
CHEAT & 0.76* & 0.84* & 0.75* & 0.82* & 0.61 & 0.98 & 0.52 & 0.93 & \textbf{0.82} & \textbf{0.99} \\
GPA & 0.83 & 0.92 & 0.86 & 0.99 & 0.62 & 0.88 & 0.52 & 0.96 & \textbf{0.99} & \textbf{1.00} \\
Sa & 0.81 & 0.95 & \textbf{0.94} & \textbf{1.00} & 0.62 & 0.97 & 0.53 & 0.92 & 0.87 & 0.996 \\
\hline
\end{tabular}
\label{tab:Polarized}
\end{table*}
\subsection{Regressor versus Classifiers}
\begin{table*}[ht]
\caption{Accuracy of existing detectors and our regression detector on continuous CAS-CS dataset with different ground truths (BST)}
\centering
\begin{tabular}{lccccccccc}
\hline
\textbf{Model/BST} & \textbf{0.1} & \textbf{0.2} & \textbf{0.3} & \textbf{0.4} & \textbf{0.5} & \textbf{0.6} & \textbf{0.7} & \textbf{0.8} & \textbf{0.9} \\
\hline
OpenAI detector & 0.63 & 0.64 & 0.68 & 0.69 & 0.68 & 0.66 & 0.64 & 0.63 & 0.64 \\
ChatGPT detector & 0.66 & 0.69 & 0.71 & 0.72 & 0.72 & 0.69 & 0.66 & 0.65 & 0.66 \\
Academic detector & 0.50 & 0.50 & 0.51 & 0.51 & 0.51 & 0.51 & 0.52 & 0.55 & 0.60 \\
DetectGPT detector & 0.58 & 0.52 & 0.52 & 0.53 & 0.51 & 0.51 & 0.52 & 0.52 & 0.52 \\
\textbf{Regression (Ours)} & \textbf{0.95} & \textbf{0.75} & \textbf{0.93} & \textbf{0.96} & \textbf{0.92} & \textbf{0.84} & \textbf{0.77} & \textbf{0.80} & \textbf{0.92} \\
\hline
\end{tabular}
\label{tab:BST}
\end{table*}

We used the BST method to evaluate the performance of each detector under various instructor requirements (as mentioned in Section \ref{sec:BST}). The results presented in Table~\ref{tab:BST} indicate that using our continuous dataset negatively affects the performance of existing detectors. Regardless of what ground truth we use to divide the CAS-CS dataset, the accuracy rate is insufficient. This finding highlights the inadequacy of binary classification detectors for identifying generated text with varying levels of human involvement.

To further demonstrate the limitations of treating the detection task as a binary classification problem, we conducted an additional experiment. Given that most existing detectors utilize a similar RoBERTa-based backbone, achieving different results primarily through fine-tuning on different datasets, we adopted the same RoBERTa-based framework. Based on the BST method mentioned before, we polarize the continuous dataset into different binary classification datasets with different thresholds (BSTs), which served as the ground truths for fine-tuning RoBERTa. Each classifier was fine-tuned on a subset of the continuous dataset, segmented by the i\% BST, resulting in classifiers we call i\% RC (retrained classifiers). We then evaluated the accuracy of each retrained classifier against their respective ground truths. 

\begin{table}[h]
\caption{Accuracy for fair comparison with classifier retrained on datasets partitioned with different ground truths (RC) and regressor with different ground truths (BSTs)}.
\centering
\begin{tabular}{lcccc}
\hline
\textbf{Model/BST} & \textbf{0.5} & \textbf{0.4} & \textbf{0.3} & \textbf{0.6} \\
\hline
0.5 RC & 0.91 & - & - & - \\
0.4 RC & - & 0.94 & - & - \\
0.3 RC & - & - & 0.91 & - \\
0.6 RC & - & - & - & 0.74 \\
\textbf{Regressor (Ours)} & \textbf{0.92} & \textbf{0.96} & \textbf{0.93} & \textbf{0.84} \\
\hline
\end{tabular}
\label{tab:fair}
\end{table}

The results (Table~\ref{tab:fair}) suggest that in real-world scenarios in which human involvement is continuous rather than discrete, the binary classification approach is inherently flawed. This experiment demonstrated that regardless of the criteria used to segment the dataset, the binary classification framework fails to accurately capture the nuanced spectrum of human involvement. These findings underscore the necessity for a regression-based approach, which better accommodates the continuous nature of human involvement in AI-generated texts. 

\subsection{Evaluation on Continuous Datasets}
We used our regressor to test the MSE of CAS-CS datasets generated by different models for two benchmarks. The results are shown in Table~\ref{tab:accuracy}. To observe the regression ability of the models more directly, we drew a scatter plot with horizontal coordinate \(y\) and vertical coordinate \(\hat{y}\) and drew a fitted line using the least squares method with the standard line \(\hat{y} = y\) as a reference. The result is shown in the Supplementary Material section. Our results show that although our model training dataset contains only the text generated by ChatGPT, it also has some generalization ability for other models, and the generalization effect was perfect for the GPT series.

\begin{table}[h]
\caption{Accuracy (±0.15) on continuous CAS-CS datasets generated by different models for MSE and ACC.}
\centering
\begin{tabular}{lcccc}
\toprule
\textbf{} & \makebox[1cm][c]{\textbf{Training set}} & \makebox[1.5cm][c]{\textbf{Testing set}} & \makebox[1.2cm][c]{\textbf{GPT-4}} & \makebox[1.2cm][c]{\textbf{Claude-3}} \\
 & (ChatGPT) & (ChatGPT) & & \\
\midrule
\textbf{MSE} & 0.004 & 0.0065 & 0.009 & 0.034 \\
\textbf{ACC} & 99.7\% & 98.3\% & 96.9\% & 69.7\% \\
\bottomrule
\end{tabular}
\vspace{0.5cm}
\begin{tabular}{lcccc}
\toprule
\textbf{} & \makebox[1.5cm][c]{\textbf{Gemini}} & \makebox[1.5cm][c]{\textbf{Falcon-7B}} & \makebox[1.2cm][c]{\textbf{GPA}} & \makebox[1.2cm][c]{\textbf{Sa}} \\
 & (Bard) & & & \\
\midrule
\textbf{MSE} & 0.025 & 0.02 & 0.0073 & 0.03 \\
\textbf{ACC} & 78.3\% & 83.1\% & 97.3\% & 68\% \\
\bottomrule
\end{tabular}
\label{tab:accuracy}
\end{table}

\subsection{Evaluation on Diverse Prompt Templates}
To address the variability in how individuals might use LLMs to generate content, we conducted an additional experiment. We recognized that users may not follow a uniform approach in creating prompts, which could affect the results. To explore this, we tested our method with several different prompt templates to assess its robustness and effectiveness across various scenarios. Detailed results can be found in Table \ref{table:template} in the supplementary material. In addition to our direct generation method, we tried three additional generation methods: "expressing a need in a student's situation," "having an LLM revise the initial generated text," and "taking five relevant articles and summarizing them into one article." The results indicate that our model maintains high accuracy and robustness across different prompt templates.

\section{Conclusion}
In this study, we developed a method for measuring human involvement in generated texts and applied it to the scenario of academic dishonesty facilitated by large language models. We presented a dual-head model based on RoBERTa for estimating both human involvement and human-provided words in the generated text. Our analysis highlighted the limitations of existing binary classification methods in detecting AI-generated content and demonstrated the advantages and flexibility of a regression-based approach. Such an approach provides instructors with accurate assessments in real-world scenarios.

Our study has several limitations. The training and validation sets are focused on computer science. While there is an ability to generalize to other fields, we did not systematically create datasets for testing in other fields. Furthermore, RoBERTa has an upper limit on text length, preventing it from processing long texts, such as full papers. The current measurement of human involvement relies on a model-based score. Future research could explore more intuitive quantification methods, such as incorporating survey data.

\section*{Acknowledgements}
This work was partially supported by JSPS KAKENHI Grants JP21H04907 and JP24H00732, by JST CREST Grant JPMJCR20D3 including AIP challenge program, by JST AIP Acceleration Grant JPMJCR24U3, and by JST K Program Grant JPMJKP24C2 Japan.
\bibliographystyle{IEEEtran}
\bibliography{Conference-LaTeX-template_10-17-19/IJCNN25}
\section{Supplementary Material}
\subsection{More Example Application Scenarios}
\begin{figure}[htbp]
\centering
\includegraphics[width=\linewidth]{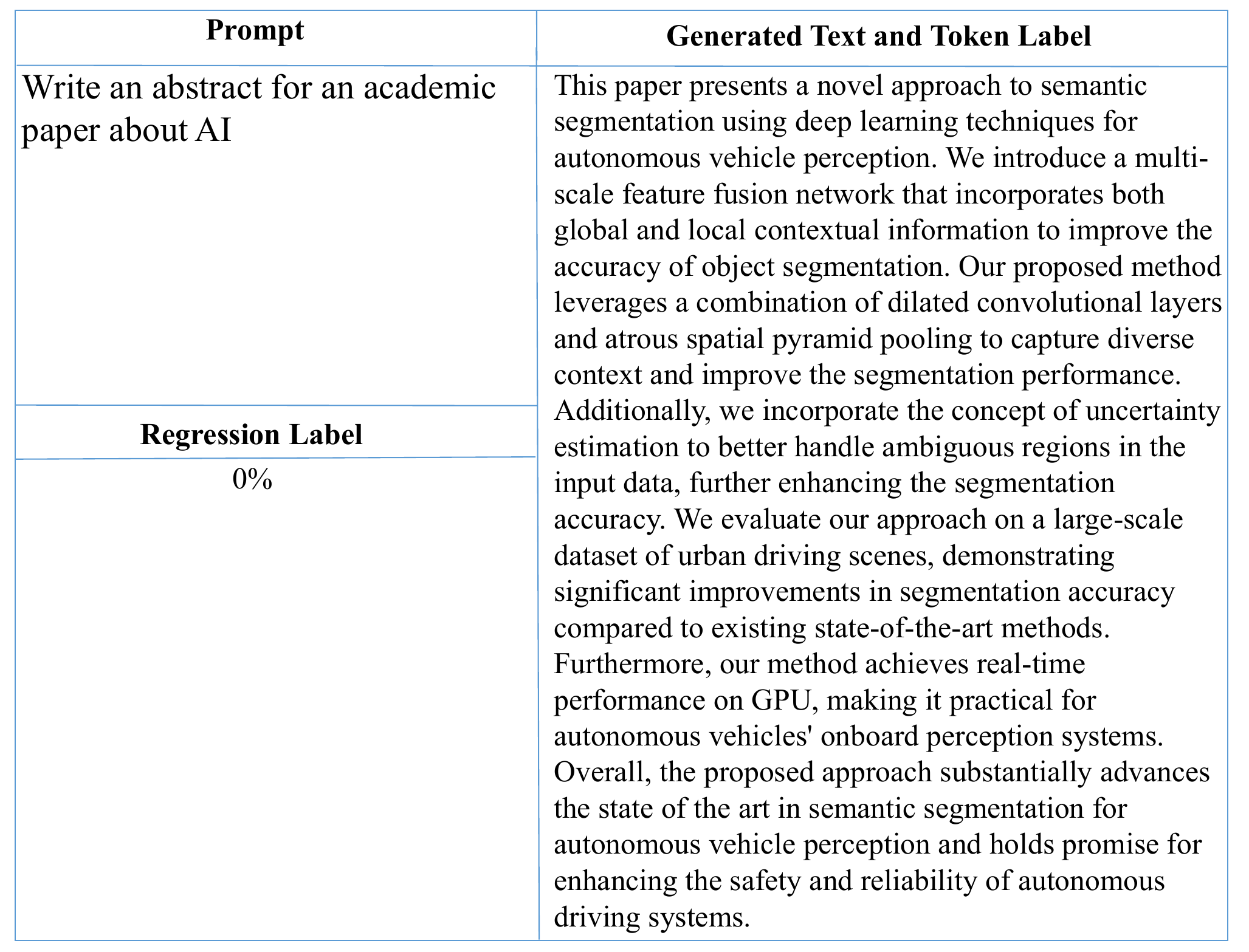} 
\caption{Abstract entirely written by ChatGPT.}
\label{fig:example1}
\end{figure}

\begin{figure}[htbp]
\centering
\includegraphics[width=\linewidth]{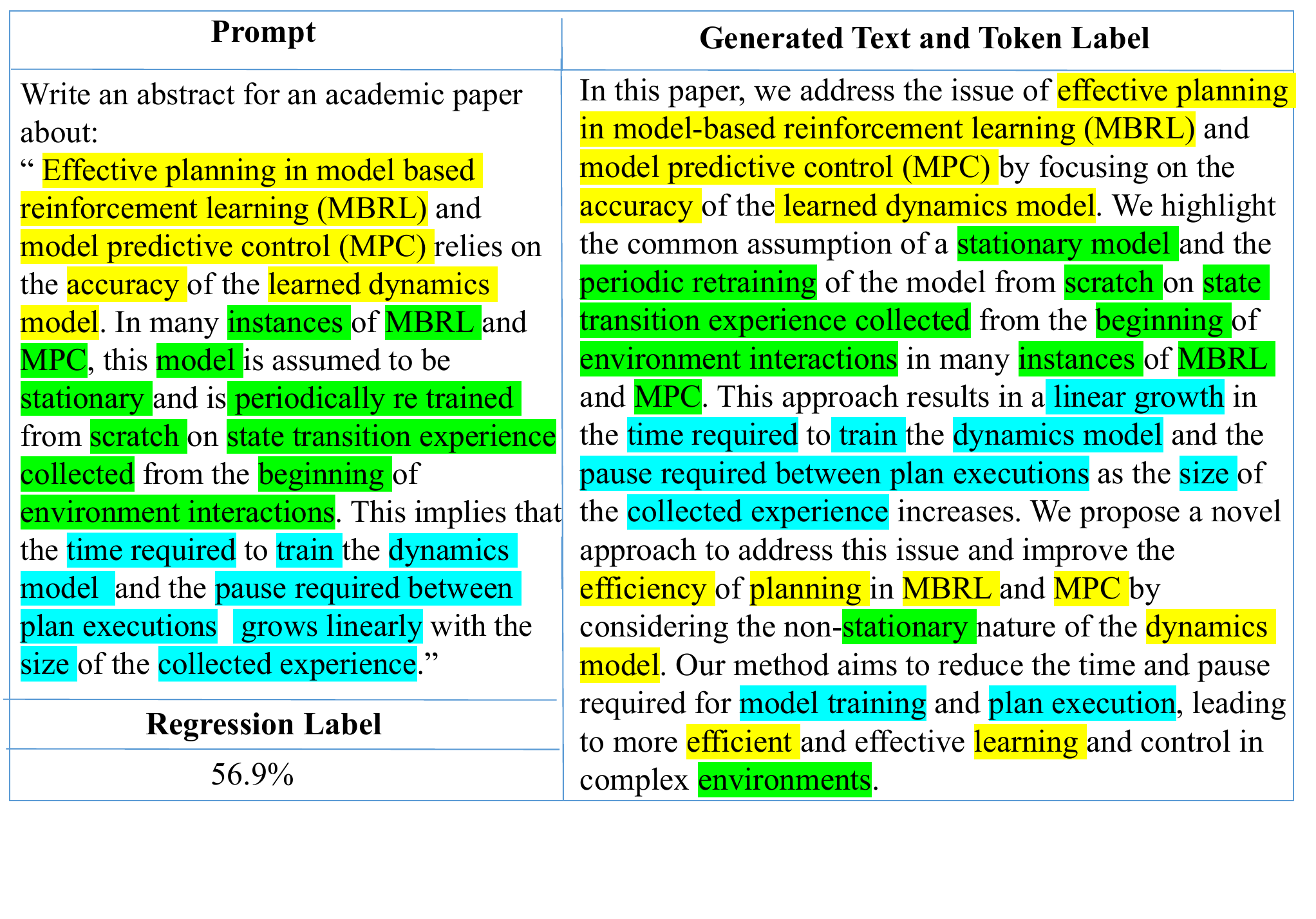} 
\caption{Abstract written on basis of human-informed prompt.}
\label{fig:example2}
\end{figure}

\begin{figure}[htbp]
\centering
\includegraphics[width=\linewidth]{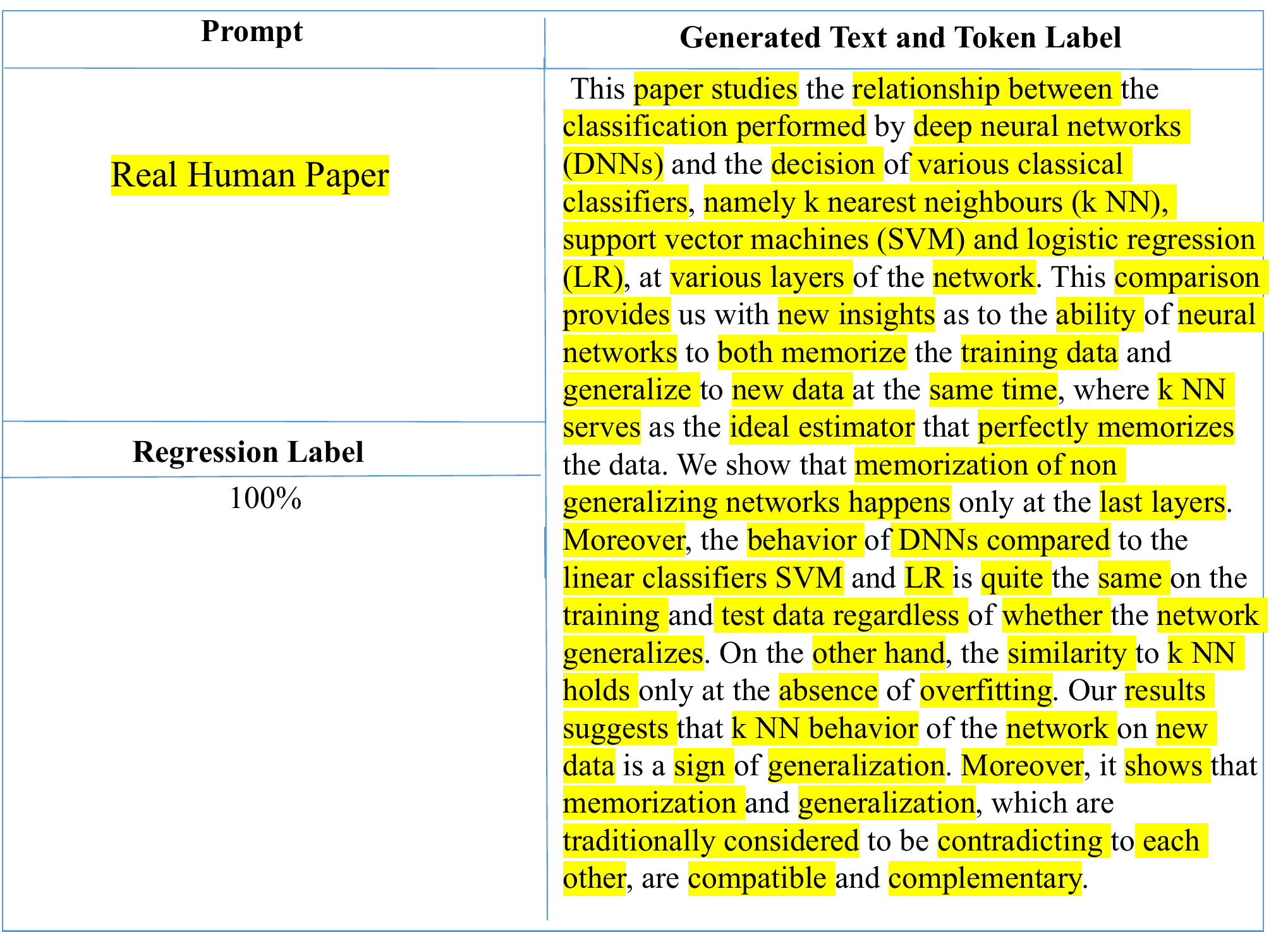} 
\caption{Abstract written by a person.}
\label{fig:example3}
\end{figure}

Figures \ref{fig:example1}, \ref{fig:example2}, and\ref{fig:example3} provide three more examples of our application scenario. The examples contain human involvement from 0\% to 100\%; the marked tokens are shown on the right.

\subsection{Introduction of BERTScore}
Unlike traditional metrics such as BLEU and ROUGE, which rely on exact n-gram matches, BERTScore leverages the contextual embeddings provided by the BERT (bidirectional encoder representations from transformers) model to capture semantic similarity between the generated and reference texts. This approach allows for a more nuanced evaluation of generated text, considering the meaning and context rather than only surface-level matches. The complete score matches each token in x to a token in \(\hat{x}\) to compute recall and each token in \(\hat{x}\) to a token in x to compute precision. For reference x and candidate \(\hat{x}\), the recall, precision, and F1 scores are
\begin{align}
R_{\text{BERT}} &= \frac{1}{|x|} \sum_{x_i \in x} \max_{\hat{x}_j \in \hat{x}} x_i^\top \hat{x}_j , \\
P_{\text{BERT}} &= \frac{1}{|\hat{x}|} \sum_{\hat{x}_j \in \hat{x}} \max_{x_i \in x} x_i^\top \hat{x}_j , \\
F_{\text{BERT}} &= 2 \frac{P_{\text{BERT}} \cdot R_{\text{BERT}}}{P_{\text{BERT}} + R_{\text{BERT}}} .
\end{align}

\subsection{Further Evaluation of Regression Ability}
\begin{figure}[htbp]
\centering
\includegraphics[width=\linewidth]{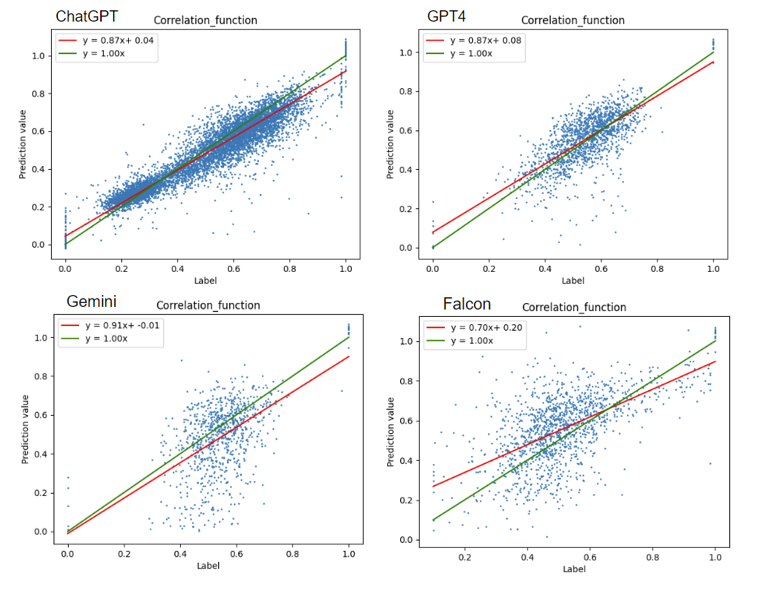} 
\caption{Result of regression with ChatGPT (upper left), GPT4 (upper right), Claude-3 (lower left), and Falcon (lower right) models. The horizontal coordinate is the true label; the vertical coordinate is the estimated value. The green line represents the best solution \(\hat{y} = y\) for reference. The red line is the fitted line.}
\label{fig:fit}
\end{figure}
To evaluate the regression ability of our regression model more directly, we drew a scatter plot with horizontal coordinate \(y\) and vertical coordinate \(\hat{y}\) and drew a least squares fitted line with standard line \(\hat{y} = y\) as a reference. The results are shown in Figure \ref{fig:fit}. 

\subsection{Evaluation using Diverse Prompt Templates}
We recognize that users may not follow a uniform approach in creating prompts, which could affect the results. To explore this effect, we tested our method using several different prompt templates to assess its robustness and effectiveness across various scenarios. Following Dou et al. \cite{dou2024enhancing}, we first split the prompt into two parts: the template summarizing the task and the content containing specific human-defined details, denoted as X. The content includes practical text, like the title of an actual paper written by actual people, alongside the task description.

For the template, the prompts for the same task can be varied in accordance with how the task is expressed. Template variants include
\begin{itemize}
\item \textbf{Direct expression}: Write an academic abstract about X.
\item \textbf{Student perspective}: I am a college student and have already finished part of an abstract about X. Please complete the abstract.
\item \textbf{Dual generation}: Revise X and then write an abstract based on the revised text.
\item \textbf{Summarization generation}: Find five abstracts about X and write a new academic abstract.
\end{itemize}

We used our detection system to calculate the MSE for various data sets on the basis of these four templates. The results are shown in Table \ref{table:template}.
\begin{table}[h]
\caption{MSE and accuracy within 15\% of true values for four templates.}
\centering
\label{table:template}
\begin{tabular}{lcccc}
\hline
\textbf{Template} & \textbf{Direct} & \textbf{Student} & \textbf{Dual} & \textbf{Summarization}\\
\hline
MSE & 0.0065 & 0.0074 & 0.0122 & 0.00857 \\
ACC & 0.983 & 0.977 & 0.896 & 0.970 \\

\hline
\end{tabular}
\end{table}

The results show that our model is robust and can deal with the detection of different input conditions to a certain extent.

\clearpage
\newpage

\end{document}